\documentclass[sigconf]{acmart}

\setcopyright{none}
\renewcommand\footnotetextcopyrightpermission[1]{}
\usepackage{booktabs}
\usepackage{graphicx}
\usepackage{amsmath}
\usepackage{multirow}
\usepackage{enumitem}
\usepackage{tikz}
\usepackage{pgfplots}
\usepackage{xcolor}
\usepackage{url}
\pgfplotsset{compat=1.18}
\usetikzlibrary{shapes.geometric, arrows.meta, positioning, fit, backgrounds, calc}

\definecolor{bestgreen}{RGB}{0,128,0}
\definecolor{regred}{RGB}{180,0,0}
\newcommand{\best}[1]{\textcolor{bestgreen}{\textbf{#1}}}

\AtBeginDocument{%
  \providecommand\BibTeX{{%
    \normalfont B\kern-0.5em{\scshape i\kern-0.25em b}\kern-0.8em\TeX}}}

\begin{document}

\title{Multi-Modal Semantic Expansion with Constrained LLM Reranking for Conversational Music Recommendation}

\author{Naman Garg}
\affiliation{%
  \institution{National Institute of Technology Kurukshetra}
  \country{India}
}
\email{hellonamangarg@gmail.com}

\author{Sarika Jain}
\affiliation{%
  \institution{National Institute of Technology Kurukshetra}
  \country{India}
}
\email{jasarika@nitkkr.ac.in}

\author{George Fazekas}
\affiliation{%
  \institution{Queen Mary University of London}
  \country{United Kingdom}
}
\email{george.fazekas@qmul.ac.uk}

\begin{abstract}
We present Team Semiintelligencn's solution for the ACM RecSys 2026 TalkPlayData Challenge, addressing conversational music recommendation through a multi-modal and personalized conversational recommender system. Our \emph{submitted} system employs a three-stage pipeline: (1)~multi-modal retrieval constructing decay-weighted centroids across seven dense embedding spaces---track- and user-level CF-BPR, Qwen3 (metadata, lyrics, attributes), CLAP audio, and SigLIP visual---supplemented by BM25 lexical retrieval and an artist substring-match signal, all fused via weighted Reciprocal Rank Fusion (RRF) with optimized signal weights; (2)~lightweight reranking (history filtering, popularity smoothing, and catalog diversity penalization); and (3)~persona-diversified response generation using GPT-4o-mini. Beyond this submitted configuration, we report development-time experiments with additional components---constrained LLM-guided artist injection, album continuation signals, XGBoost LambdaMART, and a superior GPT-4.1 response prompt---that were not deployed to Blind~B due to cost and complexity constraints. We optimize RRF weights on a 500-session development split via differential evolution~\cite{storn1997de}, improving MRR by $+19.5\%$. On Blind~A, we observe that unconstrained LLM-guided injection across 54 sessions causes catastrophic nDCG regression ($-18.9\%$), while conservative injection on only 9 sessions yields the best observed Blind~A nDCG---a finding we present as a Blind~A observation warranting further validation. The submitted system achieves a Blind~B composite score of 0.3213.
\end{abstract}

\maketitle
\pagestyle{plain}

% ============================================================================
\section{Introduction}
% ============================================================================

Conversational recommender systems (CRS) present a unique challenge at the intersection of information retrieval, natural language understanding, conversational AI, and recommender systems~\cite{gao2021advances,friedman2023leveraging}. Unlike traditional recommendation paradigms~\cite{ricci2022handbook} where static user profiles or implicit feedback logs dominate, a CRS must interpret evolving user preferences through multi-turn dialogue, retrieve relevant items from a large catalog, and simultaneously generate coherent textual responses that align with user intent~\cite{radlinski2022subjective}. This makes conversational recommendation particularly relevant to personalized recommendation, dialogue-based recommendation, and large language model (LLM)-based recommender systems.

The ACM RecSys 2026 TalkPlayData Challenge~\cite{recsys2026challenge} formalizes this paradigm with approximately 15,000 training sessions containing up to eight conversational turns between a user and a digital music assistant. Systems must output 20 ranked track UUIDs alongside a natural-language response, jointly evaluated via a composite metric assessing retrieval quality (nDCG@20), response quality (LLM-as-a-Judge), lexical diversity (Distinct-2 bigrams), and catalog diversity. No additional external datasets were used in our solution; all catalog metadata and pre-computed embeddings are from the official challenge data release. BM25 indices, centroids, and ranking signals were derived locally from these provided resources. The task therefore provides a practical setting for studying conversational music recommendation, multi-turn recommendation, and personalized music retrieval under a unified evaluation framework.

We identify three core challenges: (1)~\emph{multi-modal coverage}---the 47,071-track catalog requires fusing textual, acoustic, visual, and collaborative signals to handle both named-entity queries (``play something by Pink Floyd'') and abstract mood queries (``something dreamy and atmospheric''); (2)~\emph{LLM hallucination boundaries}---LLM-guided artist injection can degrade sharply beyond a narrow operating range, as we observe empirically; and (3)~\emph{composite metric fragility}---independent optimization of any single metric can degrade others, producing net-negative outcomes. These challenges are representative of modern multimodal recommendation systems, where semantic retrieval, collaborative filtering, content-based recommendation, and LLM reranking must be carefully integrated.

Our contributions are:
\begin{itemize}[leftmargin=*]
    \item A multi-modal retrieval system fusing seven dense embedding spaces, BM25 lexical retrieval, and an artist substring-match signal via weighted RRF with differential-evolution-optimized signal weights, improving MRR by $+19.5\%$ over uniform weighting (deployed in the submitted system).
    \item A Blind~A observation that conservative LLM-guided artist injection applied to the 9 Blind~A sessions satisfying the $\leq 5$-artist-track criterion yields the best observed Blind~A nDCG (0.4694, $+0.8\%$ over baseline), whereas aggressive injection across 54 sessions causes catastrophic nDCG regression ($-18.9\%$), highlighting position bias as a critical failure mode (not deployed to Blind~B).
    \item A persona-diversified response generation strategy that achieves high lexical diversity (0.821 Distinct-2) across 80 sessions (deployed in the submitted system).
    \item A transparent analysis of the gap between Blind~A and Blind~B performance, attributing it to deliberate deployment-time simplifications, with explicit caveats about cross-split comparisons.
\end{itemize}

% ============================================================================
\section{Task Formulation}
% ============================================================================

\subsection{Dataset and Evaluation}
The TalkPlayData catalog contains 47,071 tracks from 8,975 artists. Table~\ref{tab:dataset} summarizes the dataset statistics. The challenge organizers provide pre-computed embeddings across multiple modalities (Table~\ref{tab:modalities}). The training set comprises ${\sim}$15,000 sessions; we partitioned 500 of these as a held-out development set (Devset) for hyperparameter tuning. Additionally, the Blind~A split (80 sessions, released to participants) served as a participant-visible evaluation set for iterative ablation experiments. The final blind evaluation set (Blind~B, 80 sessions) was held out by the organizers.

\begin{table}[htbp]
  \centering
  \caption{Dataset Statistics}
  \label{tab:dataset}
  \begin{tabular}{@{}ll@{}}
    \toprule
    Property & Value \\
    \midrule
    Total tracks in catalog & 47,071 \\
    Total unique artists & 8,975 \\
    Training sessions & $\sim$15,000 \\
    Max turns per session & 8 \\
    Devset (held-out) & 500 sessions \\
    Blind~A / Blind~B & 80 sessions each \\
    Retrieval signals & 9 (Table~\ref{tab:modalities}) \\
    \bottomrule
  \end{tabular}
\end{table}

We observe three session typologies requiring adaptive routing: \emph{ultra-cold} (no user ID and no listening history), \emph{cold} (no user ID but some listening history from the conversation), and \emph{warm} (user ID available, enabling collaborative filtering). Our session classifier inspects the presence of a \texttt{user\_id} field and counts prior ``music'' role turns to classify each session, then routes it to the appropriate retrieval strategy (Section~\ref{sec:retrieval}). In the Blind~B set, we observed approximately 20 ultra-cold, 22 cold, and 38 warm sessions.

\subsection{Composite Scoring}
The official metric is:
\begin{equation}
  \text{Score} = 0.5 \cdot \text{nDCG@20} + 0.3 \cdot \tfrac{\text{Judge} - 1}{4} + 0.1 \cdot \text{LD} + 0.1 \cdot \text{CD}
  \label{eq:composite}
\end{equation}
where LD denotes Distinct-2 lexical diversity and CD denotes catalog diversity (fraction of unique tracks across all recommendations). Each $+1.0$ Judge increase contributes $+0.075$ to composite, making response quality mathematically equivalent to significant retrieval improvements.

% ============================================================================
\section{Proposed Pipeline}
\label{sec:pipeline}
% ============================================================================

\begin{figure*}[t]
    \centering
    \includegraphics[width=\textwidth]{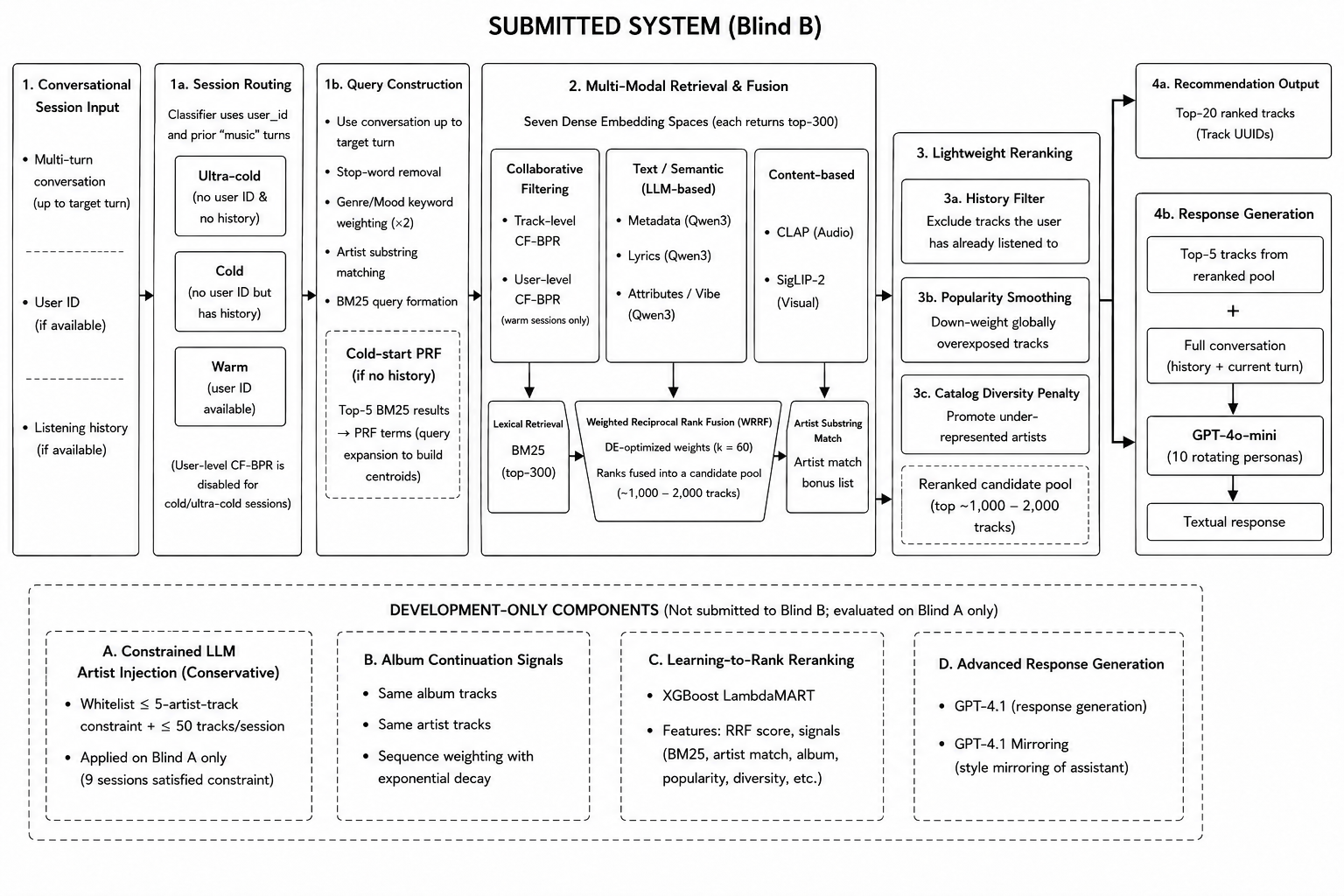}
    \caption{Overview of the submitted system (Blind B) and development-only components (evaluated on Blind A).}
    \Description{Architecture of the conversational music recommendation pipeline, showing multi-modal retrieval, reranking, and response generation, with submitted and development-only components distinguished.}
    \label{fig:pipeline}
\end{figure*}

Our pipeline consists of three stages: multi-modal retrieval (Section~\ref{sec:retrieval}), multi-signal reranking (Section~\ref{sec:rerank}), and response generation (Section~\ref{sec:response}). Table~\ref{tab:config} clarifies which components were included in the submitted Blind~B system versus those explored only during development (evaluated on Blind~A). Figure~\ref{fig:pipeline} provides an architectural overview.

\begin{table}[htbp]
  \centering
  \caption{Submitted vs.\ Development Components}
  \label{tab:config}
  \begin{tabular}{@{}lcc@{}}
    \toprule
    Component & Submitted & Dev-Only \\
    \midrule
    \multicolumn{3}{@{}l}{\emph{Stage 1: Retrieval}} \\
    \quad BM25 lexical retrieval & \checkmark & \\
    \quad 7-space dense centroid retrieval & \checkmark & \\
    \quad Optimized RRF fusion & \checkmark & \\
    \quad Artist bonus injection & \checkmark & \\
    \midrule
    \multicolumn{3}{@{}l}{\emph{Stage 2: Reranking}} \\
    \quad History filter (inline) & \checkmark & \\
    \quad Popularity smoothing & \checkmark & \\
    \quad Catalog diversity penalty & \checkmark & \\
    \quad Album continuation signal & & \checkmark \\
    \quad XGBoost LambdaMART & & \checkmark \\
    \quad Conservative GPT injection & & \checkmark \\
    \midrule
    \multicolumn{3}{@{}l}{\emph{Stage 3: Response Generation}} \\
    \quad GPT-4o-mini personas & \checkmark & \\
    \quad GPT-4.1 Mirroring prompt & & \checkmark \\
    \bottomrule
  \end{tabular}
\end{table}

\subsection{Stage 1: Multi-Modal Retrieval}
\label{sec:retrieval}

\paragraph{Query Construction and Lexical Retrieval.}
Sessions are classified as \emph{ultra-cold} (no user ID/history), \emph{cold} (no ID, some history), or \emph{warm} (ID available). Tokenized queries extract all utterances up to the target turn, removing 86 stop words (e.g., ``play,'' ``song''). To bias musically meaningful terms, curated genre (34 total, e.g., \emph{shoegaze}) and mood keywords (31 total, e.g., \emph{euphoric}) receive \emph{double-weight}. Exact substring matching ($\geq 4$ characters) against the 8,975-artist catalog detects artist mentions. We build a BM25Okapi~\cite{robertson2009bm25} index ($k_1 = 1.5$, $b = 0.75$) over concatenations of \texttt{artist\_name}, \texttt{track\_name}, \texttt{album\_name}, and \texttt{tag\_list}. Each query retrieves the top 300 candidates, serving as both direct retrievals and pseudo-relevance feedback (PRF) seeds for cold-start sessions.

\paragraph{Multi-Modal Centroids and Dialogue Elicitation.}
We utilize seven pre-computed embedding spaces, supplemented by BM25 lexical retrieval and an artist substring-match signal (Table~\ref{tab:modalities}); user-level CF-BPR is disabled for cold/ultra-cold sessions. To capture evolving multi-turn preferences (e.g., shifting from broad genres to specific acoustic traits), we construct a query centroid for each dense embedding space $j$ using the session's chronologically ordered listening history $\{h_1, \ldots, h_n\}$:
\begin{equation}
\mathbf{c}_j = \frac{\sum_{i=1}^{n} \gamma^{n-i} \cdot \phi_j(h_i)}{\left\|\sum_{i=1}^{n} \gamma^{n-i} \cdot \phi_j(h_i)\right\|_2}
\label{eq:centroid}
\end{equation}
An exponential decay factor ($\gamma = 0.85$) weights recently played tracks higher. This causes the centroid to naturally \emph{drift} toward the user's latest listening preferences while retaining accumulated listening context. For cold-start sessions lacking history, PRF from the top-5 BM25 results seeds the centroids~\cite{reimers2019sbert,karpukhin2020dpr}. Each dense embedding space yields 300 candidates ranked by cosine similarity; BM25 independently provides its top 300 lexical candidates.

\begin{table}[htbp]
  \centering
  \caption{Retrieval Signals and RRF Weights}
  \label{tab:modalities}
  \begin{tabular}{@{}llc@{}}
    \toprule
    Signal & Model / Rule & Weight \\
    \midrule
    Lexical (BM25) & BM25Okapi~\cite{robertson2009bm25} & 2.952 \\
    Collaborative (track) & CF-BPR~\cite{rendle2009bpr} & 0.874 \\
    Collaborative (user) & CF-BPR~\cite{rendle2009bpr} & 4.845 \\
    Metadata & Qwen3~\cite{qwen2025} & 0.425 \\
    Lyrics & Qwen3~\cite{qwen2025} & 0.121 \\
    Attributes/Vibe & Qwen3~\cite{qwen2025} & 1.988 \\
    Audio & LAION-CLAP~\cite{wu2023clap} & 2.066 \\
    Visual & SigLIP-2~\cite{zhai2023siglip,beyer2025siglip2} & 0.929 \\
    Artist match & Exact substring & $w_{\text{art}}=4.861$ \\
    \bottomrule
  \end{tabular}
\end{table}

\paragraph{Weighted Fusion and Optimization.}
The per-signal ranked lists are fused via weighted Reciprocal Rank Fusion (RRF) with an additive artist bonus:
\begin{equation}
\text{RRF}(d) = \sum_{j \in \mathcal{M}} \frac{w_j}{k + \text{rank}_j(d)} + \frac{w_{\text{art}}}{k+1} \cdot \mathbb{1}[\text{art}(d) \in \mathcal{A}]
\label{eq:rrf}
\end{equation}
where $k = 60$ and $\mathcal{A}$ is the set of explicitly mentioned artists. The bonus gives tracks by matched artists a rank-1-equivalent score contribution ($w_{\text{art}} = 4.861$) to ensure high recall. Signal weights $\{w_j\}$ were tuned via differential evolution~\cite{storn1997de} over a 500-session Devset, improving MRR from 0.1806 to 0.2158 ($+19.5\%$). The learned weights reveal that behavioral signals (user CF, $w=4.845$) and artist substring matches dominate, whereas lyrical content contributes minimally ($w=0.121$).

\subsection{Stage 2: Multi-Signal Reranking}
\label{sec:rerank}

The top candidates from RRF fusion pass through lightweight reranking stages. The submitted Blind~B pipeline applies three post-retrieval adjustments, all implemented inline within the scoring loop of the submission script.

\subsubsection{Submitted Reranking Stages}
The following three stages were included in the official Blind~B submission:

\textbf{1.~History Filter (inline):} Tracks appearing in the session's listening history are excluded from all retrieval and scoring operations via an inline set-membership check (\texttt{tid not in history\_set}). This enforces a non-repetition constraint ensuring our recommendations never include previously played tracks.

\textbf{2.~Popularity Smoothing:} A weak additive bonus based on challenge-provided track popularity ($s(d) \mathrel{+}= 0.05 \times \text{pop}(d) / 100$) prevents cold items from dominating when retrieval scores are near-uniform.

\textbf{3.~Catalog Diversity Penalty:} To encourage CD $> 0.03$ across all 80 evaluation sessions, tracks recommended in prior sessions receive a multiplicative penalty:
\begin{equation}
s(d) \leftarrow s(d) \times \max\!\left(0.25, \; 1 - 0.25 \, c(d)\right)
\end{equation}
where $c(d)$ is the track's prior recommendation count. This penalizes over-recommended tracks to maintain catalog diversity.

\subsubsection{Development-Only Reranking Stages (Not Submitted)}
The following components were evaluated on Blind~A (80 sessions, released to participants) but \textbf{not included} in the official Blind~B submission.

\textbf{4.~Album Continuation Signal (Dev only):} As quantified in Table~\ref{tab:album}, we discovered a powerful sequential bias: when the last 3 ground-truth (GT) tracks share an album, the next GT track shares it 58.1\% of the time (vs.\ a $\sim$15\% base rate). Adding remaining album tracks to the front of the candidate pool improved nDCG from 0.277 to 0.388 ($+40\%$) on Blind~A. It was omitted from the Blind~B submission because the album continuation pattern may reflect artifacts of the challenge data construction rather than genuine user behavior, and we could not validate this on a separate held-out set.

\textbf{5.~XGBoost LambdaMART (Dev only):} An XGBoost LambdaMART model~\cite{burges2010lambdamart} was trained on the retrieval and reranking features described above (including the album signal) and further improved nDCG from 0.388 to 0.466 ($+20\%$) on Blind~A. It was omitted from the Blind~B submission due to deployment complexity.

\textbf{6.~Conservative GPT Injection (Dev only):} GPT-4.1 was used to identify the correct artist for sessions where $\leq 5$ tracks from the target artist appeared in the top 20. When applied conservatively to only 9 such sessions, nDCG improved from 0.4659 to 0.4694 ($+0.8\%$). Under this configuration, the concurrent GPT-4.1 response prompt produced Judge~$=4.80$, LD~$=0.787$, and CD~$=0.028$, yielding a composite of $0.5 \times 0.4694 + 0.3 \times (4.80{-}1)/4 + 0.1 \times 0.787 + 0.1 \times 0.028 = 0.6012$ via Eq.~\ref{eq:composite}---the best observed Blind~A composite under the conservative-injection configuration. Notably, even stricter injection on only 3 sessions with \emph{zero} artist tracks actually underperformed the baseline ($-0.1\%$), indicating that moderate artist-track presence (1--5 tracks) is precisely where LLM intervention adds value. However, aggressive application across 54 sessions caused a catastrophic $-18.9\%$ nDCG regression ($0.4659 \!\to\! 0.3775$), consistent with severe LLM position bias when the LLM is asked to re-rank sessions where the correct artist already dominates the top ranks (Section~\ref{sec:threshold}). This component was omitted from Blind~B due to the risk that the narrow safe operating range might not generalize.

\begin{table}[htbp]
  \centering
  \caption{Album Continuation Probability}
  \label{tab:album}
  \begin{tabular}{@{}lccc@{}}
    \toprule
    Condition & Sessions & $P(\text{same album})$ & Base \\
    \midrule
    Last 3 GT share album & 487 & 58.1\% & $\sim$15\% \\
    Last 2 GT share album & 1,203 & 50.6\% & $\sim$15\% \\
    Last 1 GT from album & 3,841 & 31.2\% & $\sim$15\% \\
    \bottomrule
  \end{tabular}
\end{table}

\subsection{Stage 3: Response Generation}
\label{sec:response}

\paragraph{Persona-Driven Prompting and Constraints.}
To maximize lexical diversity (LD) and explanation quality, we generate responses using GPT-4o-mini~\cite{openai2024gpt4o} with 10 maximally differentiated personas assigned round-robin across the 80 Blind~B sessions (``passionate music blogger,'' ``veteran vinyl record store owner,'' ``late-night radio DJ,'' ``thoughtful ethnomusicologist,'' ``beat-obsessed club promoter,'' ``literary-minded music journalist,'' ``nostalgic audiophile,'' ``avant-garde curator,'' ``spoken-word poet,'' and ``synesthete experiencing music as colors''). This naturally produces distinct metaphorical registers and prevents repetitive justifications. The system prompt enforces five strict constraints: (1)~exactly 2--3 sentences; (2)~mention exactly 2 track names with their artists in natural flowing prose; (3)~briefly explain \emph{why} the tracks match the conversational context; (4)~no bullet points, numbered lists, or emojis; and (5)~no technical terms (e.g., ``algorithm,'' ``database,'' ``system,'' ``profile''). 

\paragraph{Retrieval-Augmented Generation (RAG) and Hyperparameters.}
Retrieval and generation are tightly coupled: the top-5 retrieved tracks (with title, artist, album) and full conversation log are serialized into the user prompt. This RAG framework~\cite{lewis2020rag} grounds the LLM in actual catalog items and reduces the risk of catalog hallucination, forcing it to discuss exactly 2 of the 5 retrieved tracks. To actively discourage cross-session phrase repetition and further boost the Distinct-2 bigram metric, we deploy grid-searched generation hyperparameters (also tuned on the same 500-session Devset used for RRF weight optimization): temperature $= 1.0$, max\_tokens $= 120$, top\_p $= 0.95$, presence\_penalty $= 0.6$, and frequency\_penalty $= 0.4$.

\paragraph{Development-Only Alternative (Not Submitted to Blind~B).}
On our Devset, a ``Conversational Mirroring'' prompt with GPT-4.1 achieved a substantially higher Judge score of 4.95 (vs.\ 2.60 with GPT-4o-mini personas) and LD of 0.86. Capped at $\leq 70$ words, this prompt required mirroring the user's exact vocabulary, banned 12 formulaic transitional phrases (e.g., ``Since you're into\ldots,'' ``Both tracks\ldots,'' ``Here are some\ldots''), and specified 8 varied opening patterns (e.g., \texttt{``That [quality] you love in `[track]' runs through\ldots''}). This configuration was \textbf{not included in the submitted Blind~B system} due to API cost constraints (GPT-4.1 at \$2/1M input tokens vs.\ GPT-4o-mini at \$0.15/1M). We note that its Devset advantage may not fully transfer to the blind set.
% ============================================================================
\section{Results and Analysis}
\label{sec:results}
% ============================================================================

\subsection{Official Results}
Table~\ref{tab:leaderboard} reports our Blind~B performance alongside top leaderboard entries.

\begin{table}[htbp]
  \centering
  \caption{RecSys 2026 Leaderboard (Blind~B)}
  \label{tab:leaderboard}
  \begin{tabular}{@{}llccccc@{}}
    \toprule
    Rank & Team & Comp. & nDCG & Judge & LD & CD \\
    \midrule
    1 & hallucinated & 0.689 & 0.619 & 4.75 & 0.957 & 0.028 \\
    2 & volart & 0.587 & 0.397 & 4.90 & 0.927 & 0.032 \\
    \textbf{37} & \textbf{semiintelligencn} & \textbf{0.321} & \textbf{0.232} & \textbf{2.60} & \textbf{0.821} & \textbf{0.031} \\
    \bottomrule
  \end{tabular}
\end{table}

\subsection{Ablation Analysis}

Table~\ref{tab:ablation} traces the cumulative contribution of each pipeline component, measured on the Blind~A evaluation set (80 sessions). Components are grouped by whether they were included in the submitted Blind~B system or explored only during development. Each row represents a successive pipeline version; the two largest incremental gains in the cumulative ablation came from modality expansion ($+41\%$) and artist substring-match bonus ($+131\%$).

\begin{table}[htbp]
  \centering
  \caption{Cumulative Component Ablation (Blind~A nDCG@20).}
  \label{tab:ablation}
  \begin{tabular}{@{}llc@{}}
    \toprule
    Component Added & nDCG & $\Delta$ \\
    \midrule
    \multicolumn{3}{@{}l}{\emph{Submitted to Blind~B:}} \\
    BM25 + CF-BPR (baseline) & 0.085 & --- \\
    + Multi-modal centroid anchoring & 0.120 & $+41\%$ \\
    + Artist substring-match bonus & 0.277 & $+131\%$ \\
    \midrule
    \multicolumn{3}{@{}l}{\emph{Development-only ($\dagger$):}} \\
    + Album continuation signal$^\dagger$ & 0.388 & $+40\%$ \\
    + XGBoost LambdaMART$^\dagger$ & 0.466 & $+20\%$ \\
    + Conservative GPT injection (9 sessions)$^\dagger$ & 0.469 & $+0.8\%$ \\
    \bottomrule
  \end{tabular}
\end{table}

Table~\ref{tab:response_ablation} compares response generation strategies. Note that each strategy was evaluated in the context of a different pipeline version and retrieval configuration, so the Composite column reflects both response \emph{and} retrieval quality at that development stage.

\begin{table}[htbp]
  \centering
  \small
  \setlength{\tabcolsep}{3pt}
  \caption{Response Generation Strategies (each evaluated under a different retrieval pipeline; not a controlled ablation).}
  \label{tab:response_ablation}
  \begin{tabular}{@{}lcccccc@{}}
    \toprule
    Strategy & Model & Judge & LD & CD & nDCG & Comp. \\
    \midrule
    Heuristic template & --- & 1.45 & 0.60 & 0.038$^\ddagger$ & 0.085 & 0.14 \\
    GPT-4.1 Encyclopedic & GPT-4.1 & 4.50 & 0.86 & 0.028 & 0.401 & 0.55 \\
    \best{GPT-4.1 Mirroring} & \best{GPT-4.1} & \best{4.95} & \best{0.86} & \best{0.029} & \best{0.514} & \best{0.64} \\
    GPT-4o-mini Personas$^\dagger$ & GPT-4o-mini & 2.60 & 0.82 & 0.032 & 0.232 & 0.32 \\
    \bottomrule
  \end{tabular}
\end{table}

\subsection{Error Analysis}

\textbf{Important caveat:} The highest observed Blind~A composite across all evaluated configurations was 0.64 (from the GPT-4.1 Mirroring configuration in Table~\ref{tab:response_ablation}: nDCG~$=0.514$, Judge~$=4.95$, LD~$=0.86$, CD~$=0.029$). Because these are \emph{different evaluation splits} with potentially different session distributions, the comparison provides an \emph{approximate estimate} of how deployment simplifications affected performance, not a controlled ablation.

Subject to this caveat, we attribute the gap to two deliberate deployment simplifications:

\begin{enumerate}[leftmargin=*]
  \item \textbf{Response-generation configuration change}: GPT-4o-mini with rotating personas replaced the Blind~A-validated GPT-4.1 ``Conversational Mirroring'' prompt due to cost constraints. The GPT-4.1 Mirroring configuration achieved Judge~$=4.95$, whereas the submitted GPT-4o-mini Personas configuration achieved Judge~$\approx 2.60$ under its respective pipeline; via Eq.~\ref{eq:composite}, this difference corresponds to an estimated $-0.176$ composite penalty.

  \item \textbf{Reranker omission}: The album continuation, XGBoost LTR, and conservative GPT injection stages were omitted from the Blind~B pipeline due to deployment complexity. On Blind~A, the cumulative development configuration increased nDCG from 0.277 to 0.469, corresponding to an estimated retrieval penalty of $0.5 \times (0.469 - 0.277) = -0.096$.
\end{enumerate}

Together, these two simplifications account for an \emph{estimated} composite penalty of $\sim$0.272 on Blind~A. The observed Blind~B gap is $0.64 - 0.321 = 0.319$, which is \emph{larger} than this estimate, suggesting that distributional differences between Blind~A and Blind~B also contribute to the gap.
% ============================================================================
\section{Discussion}
% ============================================================================

\textbf{The Conservative Injection Principle.}
\label{sec:threshold}
On the Blind~A set, we observe a sharp sensitivity to the aggressiveness of LLM-guided artist injection. Table~\ref{tab:threshold} reports the results across different session counts selected by an artist-track-count threshold. Counterintuitively, targeting sessions with \emph{zero} artist tracks in the top 20 (3 sessions) slightly underperforms the baseline, while targeting sessions with $\leq 5$ artist tracks (9 sessions) yields the best Blind~A result---suggesting the LLM adds value precisely in the ``partial presence'' regime where the correct artist is surfaced but under-ranked. Broadening intervention beyond this regime systematically degrades performance, and at the extreme (54 sessions) nDCG regresses by $-18.9\%$. We emphasize that this is a \emph{development-set observation} on Blind~A and may not generalize. We hypothesize that the regression reflects the LLM's tendency to displace already-correct rankings when asked to re-rank candidates for sessions with good existing retrieval, but this remains an untested hypothesis.

\begin{table}[htbp]
  \centering
  \caption{LLM Injection Aggressiveness (Blind~A nDCG@20).}
  \label{tab:threshold}
  \begin{tabular}{@{}llcc@{}}
    \toprule
    Sessions Modified & Criterion & nDCG & $\Delta$ nDCG \\
    \midrule
    0 (baseline) & --- & 0.4659 & --- \\
    3 & 0 artist tracks in top-20 & 0.4653 & $-0.1\%$ \\
    \best{9} & \best{$\leq$5 artist tracks} & \best{0.4694} & \best{$+0.8\%$} \\
    17 & $\leq$15 artist tracks & 0.4428 & $-5.0\%$ \\
    54 & All applicable (unconstrained) & 0.3775 & $-18.9\%$ \\
    \bottomrule
  \end{tabular}
\end{table}

\textbf{Signal Importance.}
The optimized RRF weights indicate that behavioral signals received substantially greater fusion weight than content-based signals. User collaborative filtering ($w=4.845$) alone outweighs all three Qwen3 text modalities combined (2.534), indicating that user-CF received the largest fusion weight under our Devset optimization, highlighting the challenge of T1 cold-start sessions, which cannot exploit the user-level CF signal.

\textbf{Composite Metric Navigation.}
The composite formula (Eq.~\ref{eq:composite}) creates a fragile multi-objective landscape~\cite{chu2023moirl}. Aggressively optimizing retrieval nDCG by letting GPT-4o rewrite responses occasionally produced very poor response-quality outcomes, yielding net-negative composite scores. This trade-off motivated our separation of retrieval and generation stages (Stages 1 and 3).

\textbf{Scalability Considerations.}
Brute-force cosine similarity over 47,071 track embeddings per embedding space is tractable for the 80-session blind set (${\sim}$5 minutes on a single CPU). However, production workloads with millions of tracks require approximate nearest-neighbor search (e.g., FAISS~\cite{johnson2021faiss} with IVF indexing). While differential evolution for weight tuning is a one-time $<30$-minute cost, the primary latency bottleneck remains the GPT API call ($\sim$1--2 seconds/session), which could be mitigated via parallelization or locally-hosted models.
% ============================================================================
\section{Related Work}
% ============================================================================

\textbf{Conversational and LLM-based Recommendation.}
Conversational recommender systems (CRS) have evolved into open-ended generative interfaces managing multi-objective interactions and subjective attributes~\cite{gao2021advances, radlinski2022subjective, chu2023moirl}, frequently deploying LLMs as interactive agents~\cite{friedman2023leveraging, jin2023core, feng2023llmcrs}. While LLMs show promise as zero-shot rankers~\cite{hou2024llmrank}, they suffer from severe position bias on large candidate sets—a vulnerability consistent with our empirical findings, motivating our constrained LLM-guided artist injection.

\textbf{Multi-Modal Retrieval and Fusion.}
Modern recommendation relies on heterogeneous feature integration~\cite{ricci2022handbook}. We unify standard dense semantic matching~\cite{reimers2019sbert, karpukhin2020dpr}, cross-modal audio-text alignments (CLAP~\cite{wu2023clap}, inspired by visual-language models~\cite{radford2021clip}), visual signals (SigLIP~\cite{zhai2023siglip, beyer2025siglip2}), sparse lexical models (BM25~\cite{robertson2009bm25}), and collaborative filtering (BPR~\cite{rendle2009bpr}). Instead of relying on parameter-free baselines, these disparate spaces are fused using Reciprocal Rank Fusion (RRF)~\cite{cormack2009rrf} with weights tuned via differential evolution~\cite{storn1997de}.

\textbf{Reranking and Retrieval-Augmented Generation.}
While retrieval pipelines traditionally refine candidates via learning-to-rank~\cite{burges2010lambdamart} or cross-encoders~\cite{nogueira2019bert}, conversational outputs require natural language justifications. This aligns CRS with Retrieval-Augmented Generation (RAG)~\cite{lewis2020rag}, where retrieved entities ground LLMs (e.g., GPT-4o, Qwen~\cite{openai2024gpt4o, qwen2025}) and reduce hallucination risk by anchoring responses in actual catalog items.

\textbf{The TalkPlayData Challenge.}
The ACM RecSys 2026 TalkPlayData Challenge~\cite{recsys2026challenge} crystallizes these advancements, extending CRS to multi-turn interactions. Our multi-modal fusion and constrained generation pipeline directly addresses its core dual-objective: jointly optimizing retrieval accuracy and generative response quality.
% ============================================================================
\section{Conclusion}
% ============================================================================

We presented Team Semiintelligencn's three-stage pipeline for conversational music recommendation. The \emph{submitted} system consists of multi-modal retrieval with differential-evolution-optimized RRF fusion across seven dense embedding spaces, BM25 lexical retrieval, and an artist substring-match signal, followed by lightweight reranking (history filtering, popularity smoothing, and catalog diversity penalization), and persona-diversified response generation using GPT-4o-mini. Beyond this, we explored---but did not deploy---album continuation signals, XGBoost LambdaMART, conservative GPT-guided artist injection, and a superior GPT-4.1 response prompt. On Blind~A, we observe a sharp sensitivity to LLM injection aggressiveness: conservative injection on 9 sessions with $\leq 5$ target-artist tracks achieves the best Blind~A nDCG (0.4694, $+0.8\%$), whereas aggressive application across 54 sessions causes catastrophic nDCG regression ($-18.9\%$). This finding, while limited to a single evaluation split, suggests that LLM-based intervention requires strict scope constraints---a direction warranting further investigation with controlled validation. Our Blind~B composite of 0.321 was limited by deliberate deployment simplifications (response-generation configuration change and reranker omission), underscoring the practical importance of faithful end-to-end deployment of validated configurations.

% ============================================================================
% ===========================================================================

\end{document}